\documentclass[conference]{IEEEtran}
\IEEEoverridecommandlockouts
\usepackage{cite}
\usepackage{amsmath,amssymb,amsfonts}
\usepackage{algorithmic}
\usepackage{graphicx}
\usepackage{textcomp}
\usepackage{xcolor}
\usepackage{url}
\def\BibTeX{{\rm B\kern-.05em{\sc i\kern-.025em b}\kern-.08em
    T\kern-.1667em\lower.7ex\hbox{E}\kern-.125emX}}
\begin{document}

\title{Ensuring Safe Physical AI in Urban Mobility via Hazard-Informed Synthesized Envelopes}

\author{\IEEEauthorblockN{Alexei Odinokov}
\IEEEauthorblockA{
\textit{SafePi.ai} and\\
\textit{Xortech DOO}\\
Belgrade, Serbia \\
aodinokov@gmail.com}
~\\
\and
\IEEEauthorblockN{Rostislav Yavorskiy*}
\IEEEauthorblockA{
\textit{SafePi.ai}\\
Madrid, Spain\\
ryavorsky@gmail.com}
*Corresponding author
}

\maketitle

\begin{abstract}

As heterogeneous robotic systems deploy across diverse urban zones, maintaining safety amid complex human-robot interactions remains a critical challenge. We present a unified framework that bridges systematic hazard analysis and runtime enforcement using hazard-informed safety envelopes. Rather than treating safety as a static constraint isolated within individual software modules, we introduce a cross-layer safety transformation process spanning symbolic, spatial, and dynamic world models. We show how this representation naturally interfaces with physical AI runtime harnesses to guarantee safe urban mobility.

\end{abstract}


\section{Introduction}
The integration of robotics into urban mobility is shifting from controlled industrial environments to public spaces. Robotics is moving onto city streets, changing how people, goods, and services move, giving rise to three dominant use cases: 
\begin{itemize}
    \item sidewalk autonomous delivery robots,  small, electric, cooler-sized, equipped with cameras, GPS, and LiDAR, travel on sidewalks at pedestrian speeds to deliver groceries and takeout \cite{jiang2025impact}, 
    \item autonomous vehicles and robotaxis that are transitioning from pilot phases to scale and operate 24/7 \cite{broekman2025toward}, 
    \item autonomous road-maintenance robots that can patrol city streets at night to detect asphalt cracking and instantly seal them\cite{kulkarni2025automatic}.
\end{itemize}

Deploying these heterogeneous robotic systems alongside human agents poses severe safety, operational, and regulatory challenges. Each domain occupies a distinct urban zone and uses a unique safety paradigm to prevent physical harm and minimize conflict.

Sidewalk robots are operating on pedestrian pathways. Their safety is ensured by limiting kinetic energy through ultra-low speeds and relying on ``yield-to-all'' behavior trees \cite{dollar2019automated}.

Robotaxis are passenger-carrying vehicles that require strict multi-layered hardware and sensor redundancies (camera, LiDAR, radar) alongside deterministic safety-envelope algorithms. The systems must conform to rigorous behavioral standards, such as IEEE 2846-2022 \cite{intelligent2022ieee}.  

Road maintenance robots operate in high-risk, unstructured construction zones. Their safety relies on real-time geofencing \cite{zimbelman2017hazards}, vehicle-to-everything (V2X) communication \cite{hasan2020securing}, and active sensor-clearing mechanisms to prevent collisions with oncoming traffic or nearby human crews \cite{weidel2024five}.

Yet safety is not a static attribute of a single software component. Instead, it is a system-level feature that evolves as information and decisions flow through different layers of the system design. 

For example, a pedestrian is represented as a vulnerable road user at the symbolic layer, as a dynamic obstacle during trajectory planning, and as a moving object with position and velocity at the control layer. Although these representations differ, they all capture the same safety objective: preventing harmful interaction with the pedestrian.

In this paper, we consider safety as a cross-layer property whose representation changes with the level of abstraction, while its underlying objective remains the same. We present a unified framework that bridges systematic hazard analysis and runtime enforcement using hazard-informed {\em urban safety envelopes} that represent the range of operating conditions under which an autonomous robot or vehicle can safely interact with its environment.

\section{Urban Safety Envelope}

We believe that safe urban mobility requires a broad universal framework aimed at maintaining sufficient safety margins before dangerous situations develop. 

\begin{figure}[htbp]
\centerline{\includegraphics[width=0.35\textwidth]{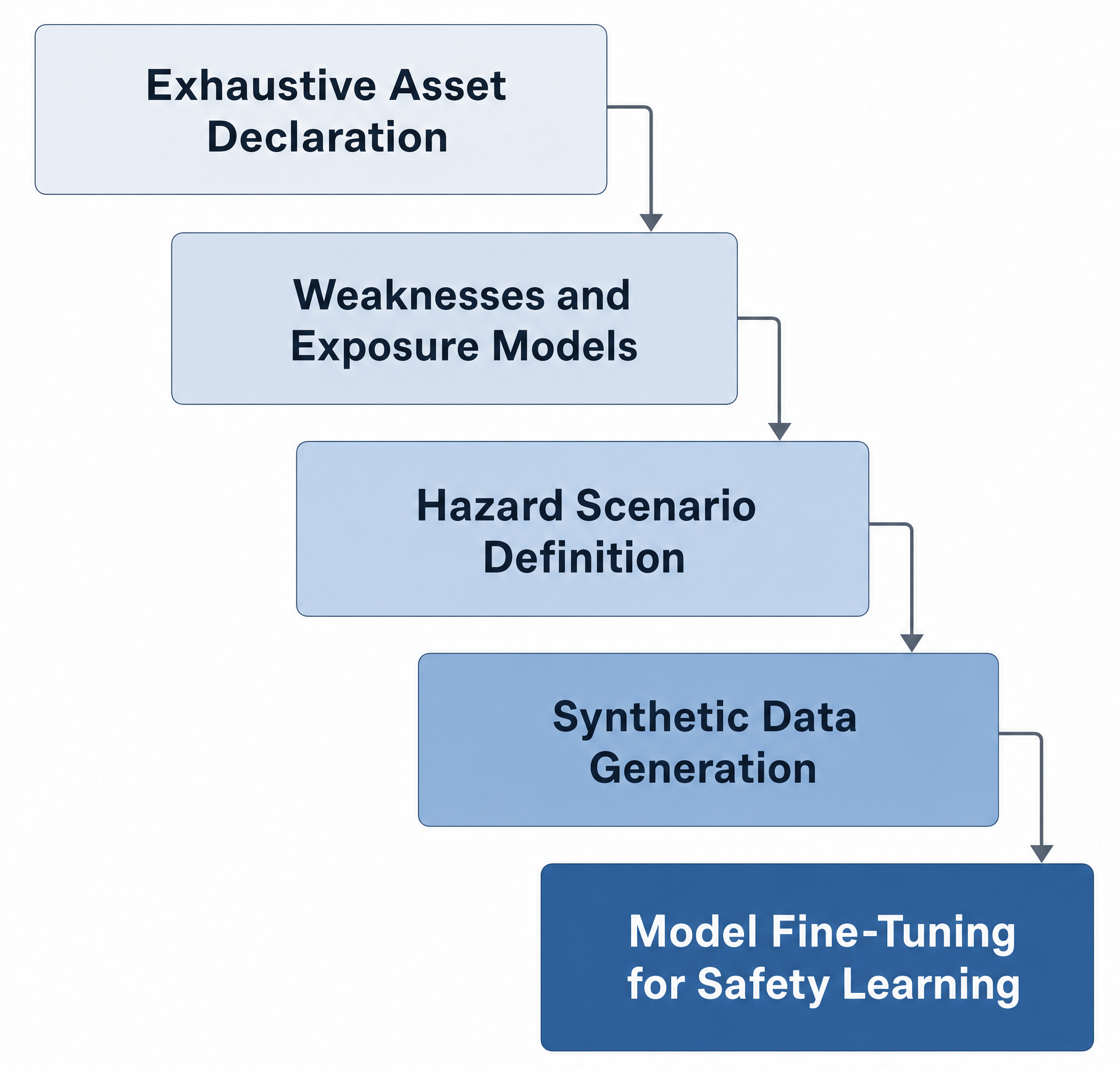}}
\caption{Hazard-informed data pipeline, see \cite{odinokov2026hazard}}
\label{fig_data_pipeline}
\end{figure}

The  urban safety envelope is learned using the hazard-informed pipeline introduced in \cite{odinokov2026hazard, dorofeev2026learning}. Rather than collecting only normal driving data, synthetic training scenarios are generated specifically around known safety-critical situations. Each scenario is derived from explicitly defined assets, exposure modes, and hazard scenarios, ensuring that the generated data focuses on conditions where safety margins are reduced, see Fig. \ref{fig_data_pipeline}.

For example, an urban intersection may include pedestrians, cyclists, parked vehicles causing visual occlusions, adverse weather, and varying traffic densities. By systematically varying these factors, the simulation generates scenarios ranging from completely safe situations to conditions that approach safety limits. This allows machine learning models to observe the gradual transition from normal operation to hazardous situations instead of learning only from rare collision events.

The resulting models continuously estimate how close the current driving situation is to violating predefined safety constraints. If the estimated safety margin becomes too small, the autonomous system can respond proactively by reducing speed, increasing following distance, selecting an alternative trajectory, or transferring control to a higher-level safety controller.

Unlike conventional hazard detection, which reacts after a dangerous situation has already emerged, the proposed approach enables predictive safety assessment. The objective is not simply to recognize hazards but to identify when the vehicle is approaching the boundary of safe operation.

An additional advantage of the proposed approach is its interpretability. Because every scenario originates from an explicitly defined hazard within the safety engineering process, the behavior learned by the model can be traced back to specific assets, vulnerabilities, and hazard scenarios identified during risk analysis. This improves transparency and provides stronger engineering evidence for safety validation and certification.

\section{Safety Across Hierarchical World Models}

Autonomous urban mobility requires reasoning across multiple levels of abstraction. No single representation of the environment is sufficient to support strategic decision making, trajectory planning, and real-time vehicle control simultaneously. Instead, autonomous systems employ a hierarchy of world models, where each layer maintains a specialized representation of the environment that captures only the information relevant to its function, see \cite{FFL}.

From a safety perspective, each layer defines its own notion of safe operation. Consequently, urban safety cannot be represented by a single model but emerges from the consistent interaction of multiple safety abstractions across the control hierarchy.


\subsection{Symbolic Safety Layer}

The highest layer represents the environment using symbolic entities, relationships, and operational rules. At this level, the system reasons about mission objectives, traffic regulations, legal constraints, and ethical requirements rather than physical geometry.

\begin{figure}[htbp]
\centerline{\includegraphics[width=0.3\textwidth]{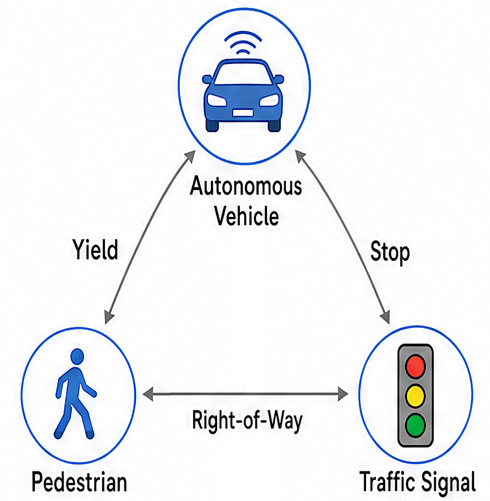}}
\caption{Symbolic safety layer}
\label{l1}
\end{figure}

Safety is expressed in terms of symbolic constraints, such as respecting pedestrian priority, obeying traffic signals, avoiding restricted areas, and complying with operational policies. The environment is therefore viewed as a structured network of permissible and prohibited actions rather than a physical space.

Within this representation, a pedestrian is treated as a vulnerable road user whose presence imposes behavioral constraints on the autonomous system.

As illustrated in the symbolic relation graph in Fig. \ref{l1}, safety rules could be visualized as directed edges (e.g., Yield, Stop) between entities.


\subsection{Spatial Safety Layer}

The intermediate layer transforms symbolic objectives into geometric motion plans. The environment is represented as a spatial map containing free space, obstacles, traversability costs, and predicted motion of surrounding traffic participants.

\begin{figure}[htbp]
\centerline{\includegraphics[width=0.3\textwidth]{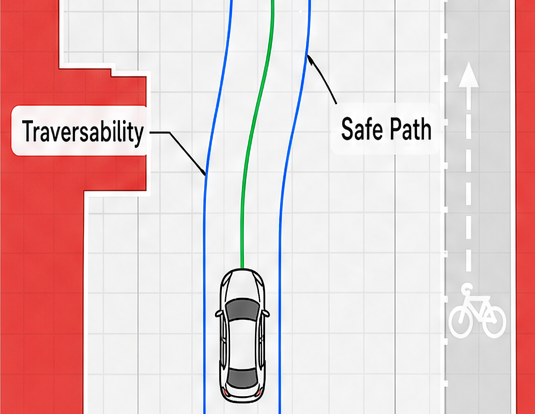}}
\caption{Spatial safety layer}
\label{l2}
\end{figure}

Safety is expressed through spatial relationships, including collision probability, minimum clearance distances, visibility constraints, and safe trajectory generation. Dynamic objects are represented by their predicted occupancy rather than by their semantic meaning.

Consequently, the pedestrian is no longer viewed as a legal entity but as a dynamic obstacle whose future motion influences the selection of a safe path.

Fig. \ref{l2} demonstrates this spatial mapping, where the continuous environment is discretized into a grid to compute a safe geometric path.


\subsection{Dynamic Safety Layer}

The lowest layer governs the physical interaction between the vehicle and its environment. Here, the world is represented by continuous dynamics, actuator limits, and physical constraints.

\begin{figure}[htbp]
\centerline{\includegraphics[width=0.48\textwidth]{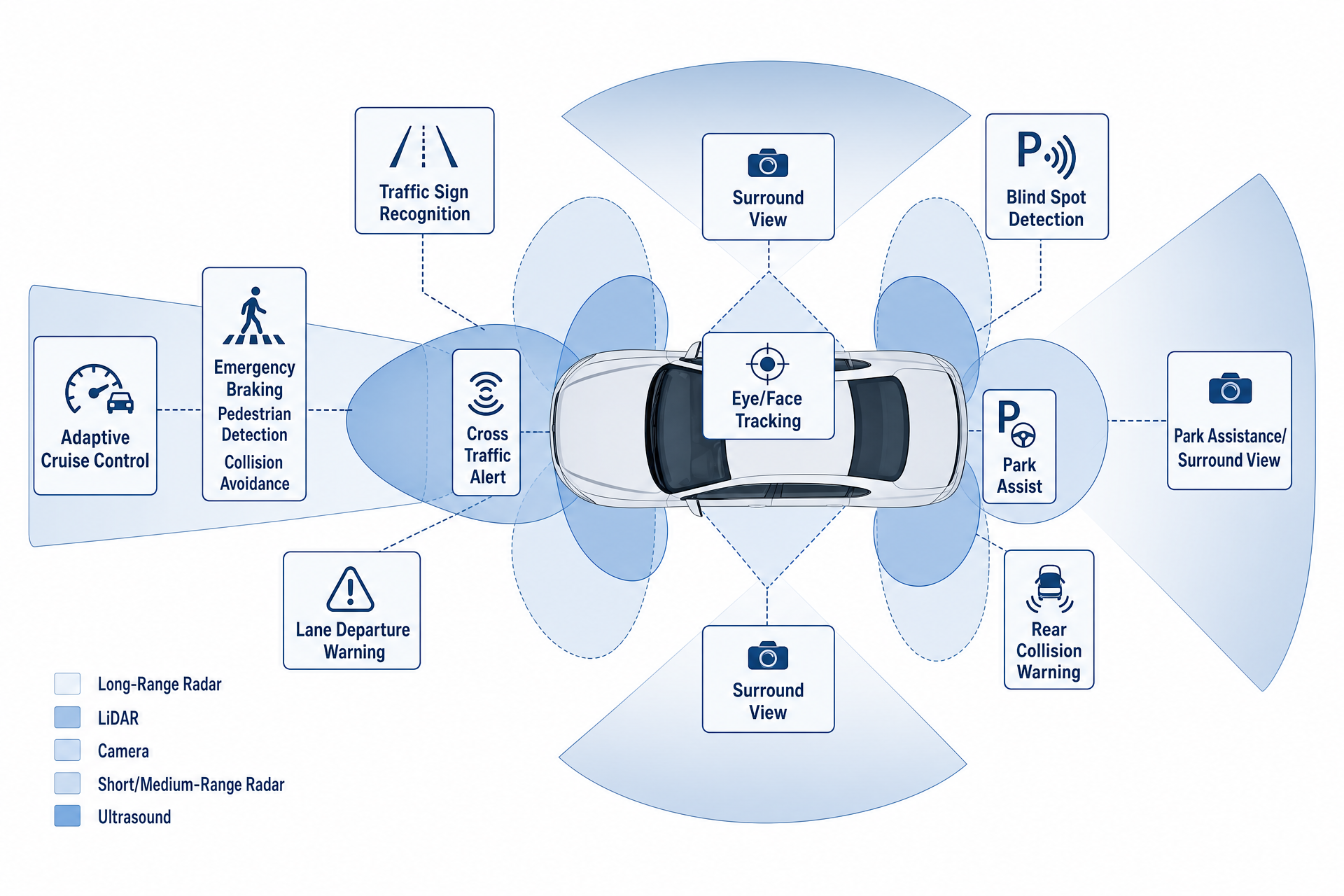}}
\caption{Dynamic safety layer}
\label{l3}
\end{figure}

Safety is defined by maintaining stable behavior while respecting limits on steering, braking, acceleration, tire-road friction, and actuator capabilities. The controller is not concerned with traffic regulations or navigation goals but with ensuring that commanded trajectories remain physically achievable.

Within this representation, a pedestrian is abstracted as a moving object characterized by position, velocity, and relative motion, enabling the controller to compute safe control actions in real time.


\section{Safety Transformation Across Layers}

The three world models described in the previous section represent the same physical environment but at different levels of abstraction. As information propagates downward through the hierarchy, safety is progressively refined from high-level symbolic objectives into low-level physical commands.
Conversely, feedback from the real world is sent upward, updating representations at each layer and enabling closed-loop adaptation. This bidirectional flow ensures that the urban safety envelope is not confined to any single layer but emerges from the consistent intersection of symbolic, spatial, and dynamic safety constraints.

\begin{figure*}[htp!]
\centerline{\includegraphics[width=0.85\textwidth]{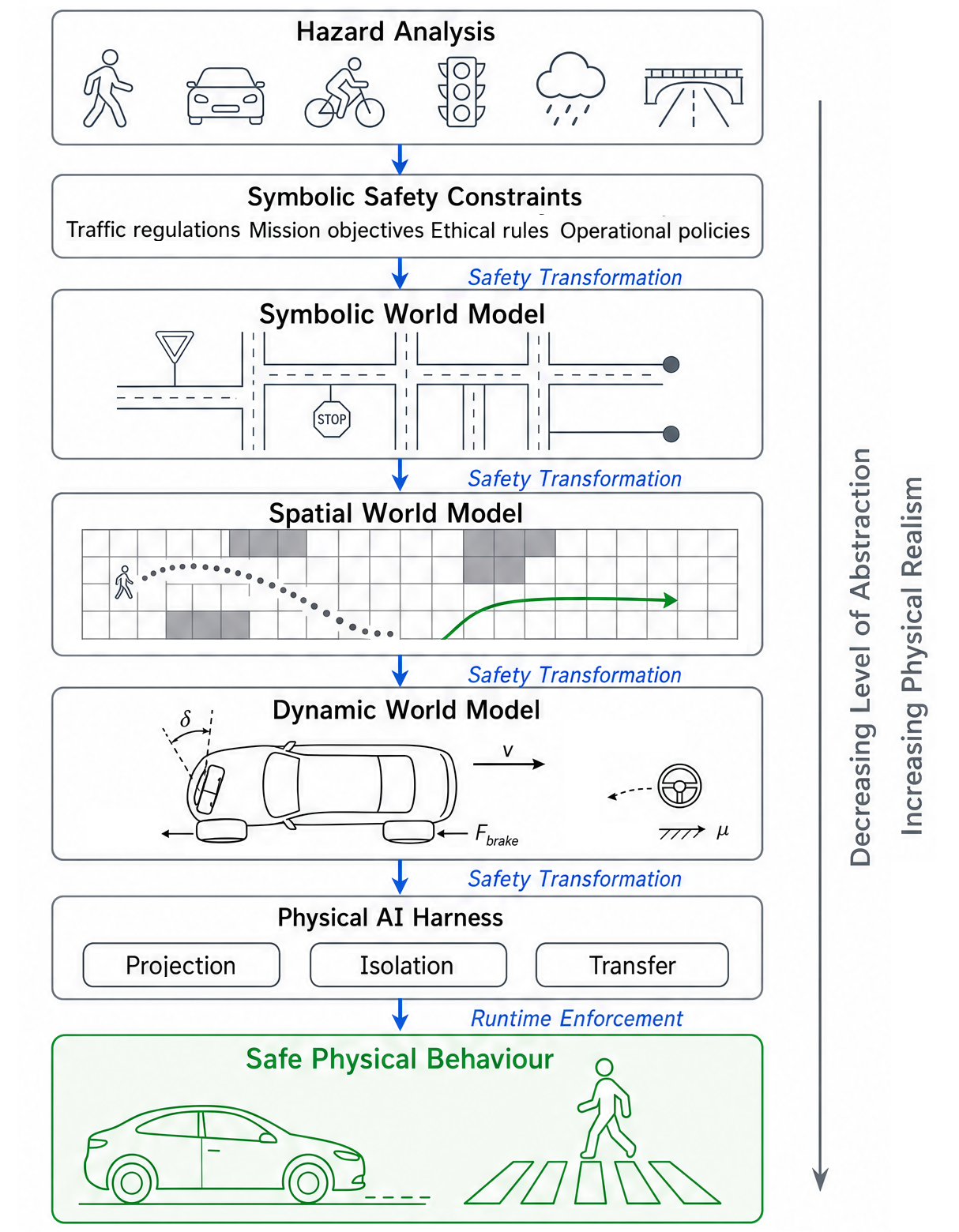}}
\caption{Safety Transformation Across Layers}
\label{fig_la}
\end{figure*}

\subsection{Forward Transformation: From Rules to Actions}

At the symbolic layer, safety is encoded as a set of mission objectives, traffic regulations, ethical principles, and operational policies. Examples include yielding to pedestrians at crosswalks, respecting speed limits, and avoiding restricted zones. These rules define permissible high-level behaviors but do not specify how to execute them geometrically or physically.

The spatial layer translates these symbolic constraints into geometric safety margins and collision-free trajectories. It reasons about the environment in terms of occupancy, clearance distances, and predicted motion of other agents. For instance, the rule ``yield to pedestrian'' is transformed into a requirement to maintain a minimum lateral distance of 0.5m and to stop before the crosswalk if the pedestrian is within 2s of arrival. The planner computes a path that satisfies these spatial margins while optimizing for efficiency and comfort.

The dynamic layer then converts the planned trajectory into physically feasible control commands: steering, braking, and acceleration—that respect actuator limits, tire‑road friction, and vehicle dynamics. At this stage, safety is expressed as maintaining stability and tracking the reference trajectory within prescribed bounds.

\subsection{Feedback Transformation: From Execution to Awareness}

Sensor measurements update the dynamic model with information about actual vehicle behavior, environmental disturbances, and unexpected obstacles. This updated physical model informs the spatial planner about deviations from the expected path and changes in the surrounding scene. If the planner detects that the original trajectory is no longer collision‑free (e.g., due to a suddenly braking vehicle), it revises the geometric plan accordingly. In turn, persistent or significant deviations may trigger a revision of symbolic decisions—for example, aborting a lane change and reverting to a safe pull‑over maneuver. This closed‑loop feedback ensures that safety remains responsive to the evolving real‑world situation.

\subsection{Emergence of the Urban Safety Envelope}

The urban safety envelope is therefore the intersection of constraints propagated through all three layers. A state is safe only if it simultaneously satisfies symbolic rules, spatial collision‑free criteria, and dynamic feasibility bounds. Learning this envelope from hazard‑informed synthetic scenarios enables the autonomous system to anticipate when it approaches the boundary of safe operation. Because the training data is systematically generated around known hazard types, the learned model captures the gradual degradation of safety margins, allowing proactive intervention before a violation occurs.

Consider a sidewalk delivery robot approaching an intersection with a pedestrian. Symbolically, the robot must yield to any crossing pedestrian. Spatially, the planner computes a path that stops before the blind corner, maintaining a buffer zone. Dynamically, the controller applies smooth deceleration within the motor’s torque limits. If the pedestrian suddenly appears closer than expected, sensor feedback updates the dynamic state, prompting the spatial planner to recalculate a tighter stop trajectory, and the controller adjusts braking force accordingly. This continuous interplay exemplifies the safety transformation across layers.

In summary, the hierarchy enables each layer to reason using the most suitable abstraction for its computational role, while the coherent propagation of safety information ensures that the overall system remains safe at all times.

\section{Runtime Safety Enforcement through the Physical AI Harness}

The previous sections describe how safety is represented and progressively transformed across multiple abstraction layers. However, preserving safety semantics during execution requires more than correct perception, planning, and control. Once machine learning models begin generating decisions in real time, an additional mechanism is required to ensure that their outputs remain consistent with the safety constraints established during system design.

Recent work by Lee et al.~\cite{lee2026harness} introduces the concept of a {\em Physical AI Harness}, a runtime software layer responsible for governing the execution of AI models within robotic systems. Unlike conventional robot middleware, which primarily coordinates communication between software components, the Physical AI Harness continuously supervises AI-generated decisions to ensure that they remain compatible with the physical capabilities of the robot and the operational requirements of its environment. 

The harness is based on three complementary mechanisms: Projection, Isolation, and Transfer (PIT), each addressing a different aspect of runtime safety.

\subsection{Projection}

Projection constrains the outputs generated by machine learning models before they are executed by the robotic system. Rather than allowing arbitrary actions, the harness projects the predicted commands onto a predefined region of acceptable behavior.

Within the proposed urban safety framework, this acceptable region corresponds naturally to the urban safety envelope. Steering commands, vehicle speed, lane changes, or trajectory modifications are accepted only if they remain within the safety margins derived from hazard analysis and safety envelope learning. Commands that violate these constraints are modified or rejected before reaching the controller.

Projection therefore represents the final validation step between learned decision making and physical execution.

\subsection{Isolation}

Modern autonomous vehicles execute numerous software components simultaneously, including perception, localization, planning, communication, and control. Machine learning models frequently compete with these components for computational resources, potentially degrading the performance of safety-critical functions.

Isolation prevents AI workloads from interfering with essential vehicle functions by allocating dedicated computational resources and enforcing execution budgets. This separation preserves the deterministic behavior required for real-time control while allowing computationally intensive AI models to operate safely alongside conventional control algorithms.

Within the urban safety framework, isolation ensures that degradation of computational performance cannot propagate into unsafe vehicle behavior.

\subsection{Transfer}

Even well-trained AI models may encounter situations outside their operational design domain. Unexpected weather, sensor failures, infrastructure damage, or previously unseen traffic scenarios may invalidate the assumptions under which the model was trained.

Transfer provides a controlled mechanism for handing authority from the AI model to a verified fallback controller whenever the system approaches unsafe operating conditions. Such fallback behavior may include emergency braking, reduced-speed operation, minimal-risk maneuvers, or transition to a conventional rule-based controller.

From the perspective of urban safety envelopes, Transfer is naturally triggered when the estimated safety margin falls below an acceptable threshold. Rather than waiting for an imminent collision, the system intervenes proactively as the vehicle approaches the boundary of safe operation.

\subsection{Relationship to the Urban Safety Envelope}

The Physical AI Harness complements the proposed hazard-informed methodology by providing runtime enforcement of the safety knowledge established during system design. Hazard analysis identifies the situations that define the urban safety envelope, safety envelope learning enables the autonomous system to estimate its proximity to unsafe operation, and the harness ensures that the resulting decisions remain physically executable and operationally safe throughout deployment.

Together, these mechanisms establish a continuous chain of safety assurance spanning hazard identification, synthetic scenario generation, machine learning, runtime monitoring, and physical execution. Safety is therefore preserved not only during model development but throughout the complete operational lifecycle of the autonomous system.

\section{Conclusion}

The proposed framework provides a unified interpretation in which safety information flows continuously through the autonomy stack, changing its representation at each abstraction level while preserving a consistent safety objective, whereas runtime enforcement ensures that the learned safety constraints remain valid during deployment.

To validate the proposed framework in practice, we are working on a proof-of-concept implementation leveraging the NVIDIA stack of world models and simulation technologies. The closed-loop validation pipeline integrates three complementary NVIDIA platforms. First, NVIDIA Isaac Sim \cite{gao2026nvidia} and Isaac Lab \cite{mittal2025isaac} serve as the primary robotics simulation and learning frameworks, providing physically accurate environments for training and evaluating autonomous agents. Second, NVIDIA Cosmos world foundation models \cite{agarwal2025cosmos} generate high-quality synthetic data and simulate diverse urban scenarios, ranging from nominal traffic conditions to safety-critical edge cases.  Third, NVIDIA Omniverse \cite{caballero2025survey} provides the underlying infrastructure for building high-fidelity digital twins of urban environments. The integration of these platforms allows us to instantiate the symbolic, spatial, and dynamic layers described in this paper within a unified simulation environment, systematically evaluate the urban safety envelope across thousands of scenario variations, and validate the PIT runtime enforcement mechanisms under controlled yet realistic conditions. Through this implementation, we aim to demonstrate that hazard-informed safety envelopes, when realized atop modern physical AI infrastructure, can provide a rigorous and scalable foundation for safe urban mobility.

\bibliographystyle{plain}
\bibliography{lit}
\end{document}